%% file: neurips_data_2024.tex
\documentclass{article}

\usepackage[nonatbib, preprint]{neurips_data_2024}

\usepackage[utf8]{inputenc} 
\usepackage[T1]{fontenc}    
\usepackage{hyperref}       
\usepackage{url}            
\usepackage{booktabs}       
\usepackage{amsfonts}       
\usepackage{nicefrac}       
\usepackage{microtype}      
\usepackage{xcolor}         
\usepackage{tcolorbox}

\usepackage{graphicx}
\usepackage{multirow}
\usepackage{makecell}
\definecolor{ForestGreen}{RGB}{34,139,34}
\definecolor{dino}{RGB}{249,231,227}
\definecolor{aliceblue}{rgb}{0.94, 0.97, 1.0}

\usepackage{diagbox} 
\usepackage{colortbl} 
\newcolumntype{b}{>{\columncolor{dino}}c}

\title{PathGen-1.6M: 1.6 Million Pathology Image-text Pairs Generation through Multi-agent Collaboration}

\makeatletter
\def\@fnsymbol#1{\ensuremath{%
		\ifcase#1
		\or 
		\dagger
		\or 
		\ddagger
		\or 
		\mathsection
		\or 
		\mathparagraph
		\else 
		\@ctrerr  
		\fi}}   
\makeatother 
\author{%
	\\
	\bf Yuxuan Sun$^{1,2}$, Yunlong Zhang$^{1,2}$, Yixuan Si$^{2}$, Chenglu Zhu$^{2}$, Zhongyi Shui$^{1,2}$,\\ 
	\bf Kai Zhang$^{3}$, Jingxiong Li$^{1,2}$, Xingheng Lyu$^{2}$, Tao Lin$^{2,}$\protect\footnotemark[1] ,
Lin Yang$^{2,}$\thanks{Corresponding author.}\\
	$^1$College of Computer Science and Technology, Zhejiang University, China \\
	$^2$Research Center for Industries of the Future and School of Engineering, Westlake University, China\\
	$^3$Department of Computer Science and Engineering, The Ohio State University, USA   
}

\begin{document}

\maketitle

\begin{figure}[h]
\vspace{-1.5em}
  \centering
  \includegraphics[width=\textwidth]{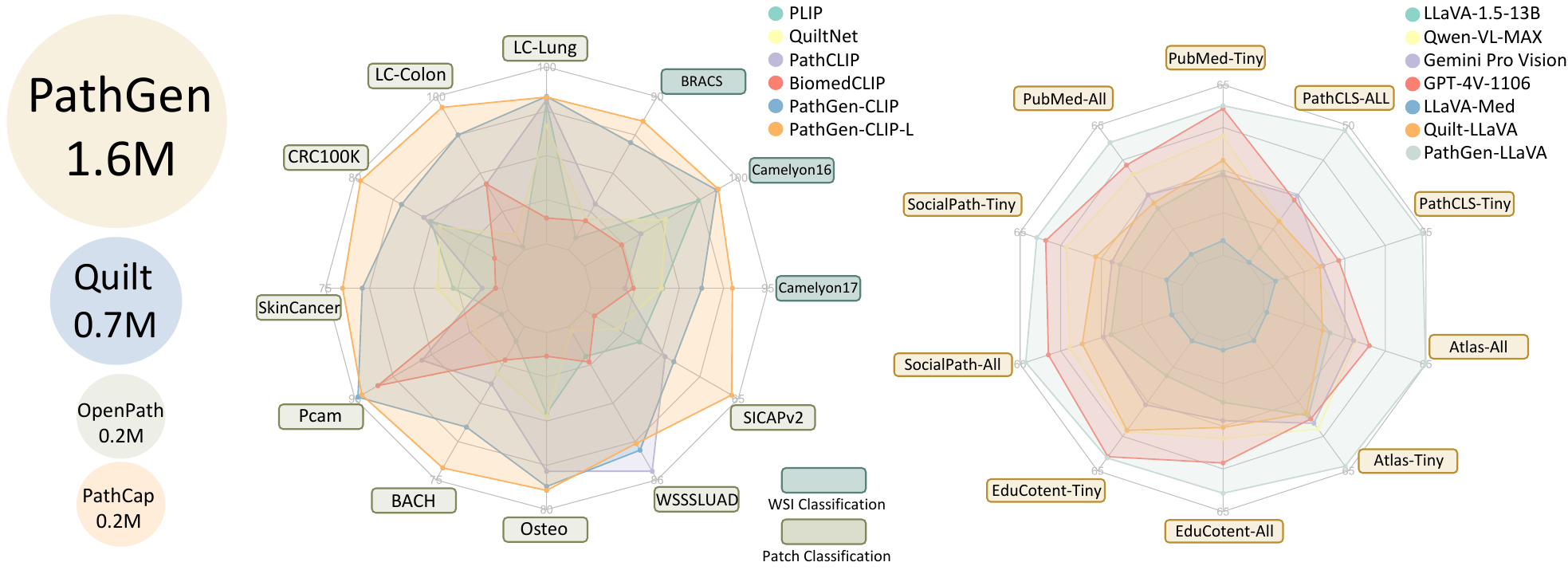}
  \caption{Illustration of the scale of the PathGen dataset (left), the performance of the proposed PathGen-CLIP (middle), and the PathGen-LLaVA (right), both derived from training on PathGen.}
  \label{fig:pre_fig}
\end{figure}
\begin{abstract}
Vision Language Models (VLMs) like CLIP have attracted substantial attention in pathology, serving as backbones for applications such as zero-shot image classification and Whole Slide Image (WSI) analysis. Additionally, they can function as vision encoders when combined with large language models (LLMs) to support broader capabilities. Current efforts to train pathology VLMs rely on pathology image-text pairs from platforms like PubMed, YouTube, and Twitter, which provide limited, unscalable data with generally suboptimal image quality. In this work, we leverage large-scale WSI datasets like TCGA to extract numerous high-quality image patches. We then train a large multimodal model to generate captions for these images, creating PathGen-1.6M, a dataset containing 1.6 million high-quality image-caption pairs. Our approach involves multiple agent models collaborating to extract representative WSI patches, generating and refining captions to obtain high-quality image-text pairs. Extensive experiments show that integrating these generated pairs with existing datasets to train a pathology-specific CLIP model, PathGen-CLIP, significantly enhances its ability to analyze pathological images, with substantial improvements across nine pathology-related zero-shot image classification tasks and three whole-slide image tasks. Furthermore, we construct 200K instruction-tuning data based on PathGen-1.6M and integrate PathGen-CLIP with the Vicuna LLM to create more powerful multimodal models through instruction tuning.
Overall, we provide a scalable pathway for high-quality data generation in pathology, paving the way for next-generation general pathology models.

\end{abstract}

\input{sec/1_introduction}
\input{sec/2_related_work}
\input{sec/3_methodology}

\input{sec/4_experiment}

\input{sec/5_conclusion}

\section{Acknowledgements}
This study was partially supported by the National Natural Science Foundation of China (Grant No.92270108), Zhejiang Provincial Natural Science Foundation of China (Grant No.XHD23F0201), and the Research Center for Industries of the Future (RCIF) at Westlake University.

\newpage
\bibliographystyle{plain}
\bibliography{main.bib}

\end{document}

%% file: sec/1_introduction.tex
\section{Introduction}

Pathology plays a crucial role in modern medicine as it is the gold standard for disease diagnosis and the selection of treatment methods \cite{kumar2014robbins}.
With the rapid growth of artificial intelligence, there is an increasing interest in developing robust general-purpose models to assist physicians, particularly in pathology. Pathology-specific CLIP models have demonstrated exceptional performance in zero-shot image classification \cite{radford2021learning,jia2021scaling}, multimodal understanding \cite{liu2024visual,liu2023improved,bai2023qwen,li2022blip,li2023blip,dai2024instructblip}, robustness to various perturbations \cite{radford2021learning,shu2023clipood,zhou2022learning}, and scalability across diverse tasks \cite{rombach2022high,lin2023clip,esmaeilpour2022zero}.

However, training such models typically requires vast amounts of data. For instance, general CLIP models are trained using massive amounts of data from sources like WIT \cite{radford2021learning} and LAION \cite{schuhmann2022laion}, at scales of millions or even billions. In the field of pathology, researchers are similarly focused on amassing large collections of pathology image-text pairs from various sources, including academic articles from PubMed \cite{lin2023pmc,ikezogwo2024quilt,sun2024pathasst}, social media \cite{huang2023visual,ikezogwo2024quilt}, and books \cite{lin2023pmc,ikezogwo2024quilt}.

Despite these efforts, the largest datasets do not exceed one million samples, which is significantly smaller compared to datasets of natural images. This underscores the challenges in the availability and scalability of pathology-related datasets. Key limitations include: (1) The pool of image-text pairs from websites and books is quickly exhausted, hindering scalability. (2) The image quality in these datasets often suffers due to compression during collection. For example, images from PubMed articles lose clarity, and educational videos on platforms like YouTube are often screenshots of presentations at 1080p resolution, further degraded by video compression, making them incomparable to the high-resolution images used in practical scenarios. (3) Unpaired image-text pairs frequently appear on social media, where users may post a pathology image but only comment on its aesthetic appeal, providing no meaningful description.

Fortunately, The Cancer Genome Atlas (TCGA) is a comprehensive, publicly funded project that provides clinical data across various cancer types. This dataset includes numerous WSIs, high-resolution scans from patient tissue samples. These WSIs inherently contain an immense amount of detailed information—such as cellular structures, tissue organization, and morphological patterns—crucial for cancer diagnosis and research. However, these datasets typically only provide labels at the slide level, leaving patches within WSIs without detailed textual annotations. This lack of specific annotations hinders models from learning rich semantic information from such high-quality image data.

In this study, we aim to harness high-quality images from WSIs to construct a large-scale image-text dataset. We develop a cascaded approach involving multiple agent models that collaborate to extract the most representative patches from WSIs and generate captions describing the visual details within each patch. This process enables us to compile a dataset of 1.6 million image-caption pairs, designed to train and enhance pathology-specific multimodal models. As depicted in Figure \ref{fig:pre_fig}, our PathGen, currently the largest pathology image-text dataset, provides high-quality data that significantly boosts the performance of existing models like CLIP and LLaVA in the pathology domain. This approach offers a scalable solution to expand the currently limited pool of pathology image-text pairs.

%% file: sec/2_related_work.tex
\section{Related work}

\textbf{Existing Vision-language Datasets.} Training vision-language models such as Contrastive Language-Image Pretraining (CLIP) \cite{radford2021learning} necessitates large and high-quality datasets of image-text pairs to capture the richness of visual and semantic information. 
In the general domain, notable datasets have been constructed in this endeavor such as LAION-5B \cite{schuhmann2022laion}, YFCC100M \cite{thomee2016yfcc100m} and  WIT-400M \cite{radford2021learning}. 
In the pathology field, the PathCap \cite{sun2024pathasst} dataset contains 207,000 pathology image-caption pairs, meticulously compiled from over 15 million image-text pairs sourced from PubMed and various textbooks. The OpenPath \cite{huang2023visual} dataset includes 208,414 pairs from Twitter posts, while the QUILT \cite{ikezogwo2024quilt} dataset gathers 768,826 histopathology image-text pairs from video frames and corresponding subtitles on YouTube. These datasets are primarily gathered from social media platforms and textbooks. Although fine-tuning CLIP on these specialized pathology datasets significantly enhances its adaptation to pathology tasks, challenges such as low image quality, image-text misalignment, and poor data scalability continue to hinder the further development of pathology-specific CLIP models.

\textbf{Pathology-specific CLIP Models.} 
CLIP \cite{radford2021learning} is a powerful model that learns visual concepts through natural language supervision. In the general domain, CLIP has demonstrated remarkable capabilities in zero-shot classification, image retrieval, and multimodal understanding by training on a vast dataset of images paired with textual descriptions. This approach leverages the rich semantic content of language to enhance visual recognition tasks, making it adaptable across various applications without the need for task-specific training. In medical and pathology imaging, CLIP's potential is being increasingly recognized as a solution to major challenges such as the scarcity of labeled data and the requirement for domain-specific expertise. By harnessing the descriptive power of natural language, CLIP aids in the identification and classification of complex pathological features, which are often difficult to annotate manually. Recently, there has been a notable increase in CLIP models specifically developed for the biomedical field, including PubMedCLIP \cite{eslami2023pubmedclip}, MedCLIP \cite{wang2022medclip}, BiomedCLIP \cite{zhang2023biomedclip}, PMC-CLIP \cite{lin2023pmc}, Quilt-Net \cite{ikezogwo2024quilt}, PathCLIP \cite{sun2024pathasst}, PLIP \cite{huang2023visual}, and CONCH \cite{lu2024visual}.

\textbf{Large Multimodal Model (LMM).}
The integration of large language models (LLMs) like T5~\cite{raffel2020exploring} and GPT-4~\cite{gpt4} with vision capabilities has spurred the development of sophisticated multimodal models (LMMs). These LMMs, such as Flamingo~\cite{alayrac2022flamingo}, BLIP-2~\cite{li2023blip}, and Fuyu~\cite{fuyu-8b}, excel in multimodal understanding by utilizing pretraining techniques. Additionally, instruction-tuning, derived from NLP, has been adapted for LMMs, enabling them to generate more controllable and task-specific outputs through datasets like those used in GPT-4V~\cite{openai2023gpt4v}, Gemini Pro Vision~\cite{team2023gemini}, Qwen-VL~\cite{bai2023qwen}, and InstructBLIP~\cite{instructblip}. 
The application of LMMs in pathology is particularly promising. Models such as PathAsst \cite{sun2024pathasst}, LLaVA-Med \cite{li2023llava},  Quilt-LLaVA \cite{seyfioglu2023quilt} have been developed using curated pathology-specific instruction-tuning datasets sourced from resources like PubMed and educational YouTube videos. These advancements facilitate the effective analysis of pathological images and the generation of descriptive texts for pathology image patches. Consequently, we leverage this capability to generate corresponding descriptions for image patches within WSIs. By creating high-quality image-text pairs, we aim to enhance the foundational vision-language models in pathology.

\textbf{Multi-Agent Collaboration.}
With the advancement of large models (LMs) and the development of specialized models for various tasks, recent research has explored the use of multi-agent collaboration. This approach allows these models to work together, achieving tasks that are beyond the capabilities of any single model alone. For instance, leveraging LLMs for role-playing can be used to accomplish tasks such as software development\cite{hong2023metagpt,qian2023communicative}, societal simulation\cite{park2023generative,park2022social}, policy simulation\cite{xiao2023simulating,hua2023war}, game simulation\cite{xu2023language,wang2023avalon} and video generation\cite{yuan2024mora}.

%% file: sec/3_methodology.tex
\section{PathGen Dataset Construction}

\begin{figure}[t]
	\centering
	\includegraphics[width=\linewidth]{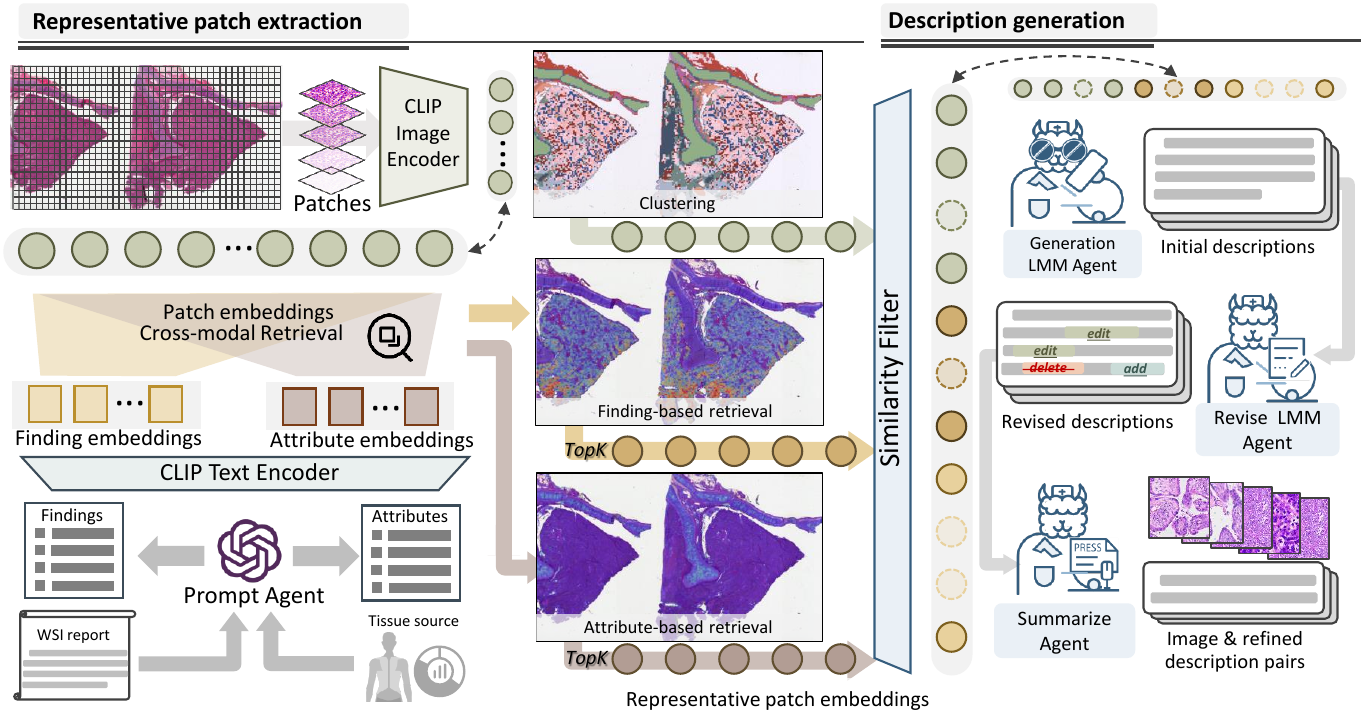}
     \vspace{-1em}
	\caption{Illustration of the multi-agent collaboration pipeline for generating pathology image-text pairs. This process comprises two main components: (1) Representative Patches Extraction, which utilizes prompt-based cross-modal retrieval and clustering; and (2) Description Generation, where multiple LMM and LLM agents are employed to generate, revise, and summarize descriptions.}
	\label{fig:framework}
    \vspace{-0.8em}
\end{figure}

The entire process of our data construction is illustrated in Figure \ref{fig:framework}. We employ multiple agents working collaboratively to generate high-quality pathology image-text pairs. This process involves extracting representative WSI image patches by generating text prompts for CLIP to retrieve the most relevant patches. These patches are then described by a trained pathology LMM agent, followed by another LMM agent that revises and summarizes the descriptions. In this section, we will introduce the construction of the agent model and explain how the agents collaborate to generate the data.

\subsection{Agent Model Preparation}
\noindent\textbf{PathGen-CLIP-L$_{init}$:} General models like OpenAI's CLIP perform suboptimally in the pathology domain, necessitating a specialized model for tasks such as pathology cross-modal retrieval we employ in section \ref{sec:detail_construction}. For this purpose, we combine portions of existing datasets, including PathCap (200K), Quilt-1M (400K), and OpenPath (100K), resulting in a total of 700K samples. We refer to this dataset as PathGen$_{init}$.
Utilizing the OpenCLIP framework\footnote{\url{https://github.com/mlfoundations/open_clip}}, we train a large version of the pathology-specific CLIP model with a 336 input image size, designated as PathGen-CLIP-L$_{init}$.

\noindent\textbf{Description LMM Agent:}
To generate superior pathology image-text pairs, we require a robust pathology-specific LMM capable of generating captions that accurately reflect the intricate details in pathology images, such as `nuclear atypia' and `disordered cellular arrangements'. Existing image-caption pairs are often too simplistic to comprehensively describe these details, hindering the training of more effective description generation models. Drawing inspiration from the PathMMU benchmark\cite{sun2024pathmmu}, we sample 10,000 image-caption pairs from PathCap, OpenPath, and Quilt-1M, respectively. We provide each image along with its corresponding caption to GPT-4V, allowing the model to enhance and refine the original captions by incorporating details observed in the images. This approach enables us to generate 30,000 detailed image descriptions.
We employ the LLaVA-v1.5-13B \cite{liu2024visual} and replace its OpenAI-CLIP vision encoder with our trained PathGen-CLIP-L$_{init}$, serving as the vision encoder for our pathology-specific description LMM agent. By training with the collected pathology image description samples, we develop the description LMM agent PathGen-LLaVA$_{desp}$, which gains robust capabilities to accurately describe pathology images. We compare the pathology image description generation capabilities of PathGen-LLaVA$_{desp}$ with those of LLaVA-Med \cite{li2023llava} and Quilt-LLaVA \cite{seyfioglu2023quilt} in supplementary materials, demonstrating its superior performance.

\noindent\textbf{Revise LMM Agent}: The Revise Agent is a pathology LMM designed with error correction capabilities. While existing pathology LMMs are trained for specific tasks like multiple-choice questions and dialogue, they lack self-correction abilities. To bridge this gap, we utilize descriptions from the Description LMM Agent and prompt GPT-4 to alter these descriptions by incorporating inaccuracies or contradictions. These alterations fall into three categories: additions, deletions, or modifications. This process generates pairs of accurate and modified descriptions, along with the specific operations that transformed the accurate descriptions into inaccurate ones. By reversing these operations—converting additions back to deletions, deletions to additions, and reversing edits—we create inverse operations that correct the inaccurate descriptions. These inverse operations, together with the descriptions before and after revise operation, are employed to train the Revise LMM Agent, thereby enhancing its capability for robust error correction.

\noindent\textbf{Summarize Agent:} Due to the CLIP model's limitation of accepting only 77 tokens as input, the data generated by PathGen-LLaVA$_{desp}$ often exceeds this length. To address this, we prompt GPT-4 to generate instruction-tuning data for summarizing these descriptions. We then fine-tune Llama-2 as a summary agent to produce concise summaries for each generated description of WSI patches.

\subsection{Details of Data Construction Pipelines}
\label{sec:detail_construction}
\noindent\textbf{Source Data:} We source approximately 7300 WSIs along with accompanying pathology reports from the TCGA dataset. These reports often contain numerous descriptions unrelated to the WSIs, such as gross findings and dimensions in centimeters. Inspired by HistGen's approach \cite{guo2024histgen} to handling reports, we refine them using GPT-4 prompts to focus more on morphological and diagnostic findings. Note that some reports contain extensive information, resulting in extracted findings that exceed the input length limit for CLIP. To address this, we design GPT-4 prompts (detailed in supplementary materials) to divide longer findings into multiple sentences. This refinement aims to facilitate the extraction of representative image patches for subsequent analysis. Based on these WSIs, we design a meticulous process from representative patch extraction to image description generation, which includes five steps outlined below to generate high-quality image-text pairs.

\noindent\textbf{Step1: Representative Patch Extraction:} In this step, we employ a prompt-based retrieval and k-means clustering to identify the most representative image patches from WSIs. \textbf{\textit{(1) Prompt-based retrieval}}: We generate specific prompts for each WSI and utilize PathGen-CLIP$_{init}$-L to retrieve relevant patches. The prompt generation can be performed in two ways: The first method involves using the previously extracted pathology findings from the report as prompts. The second way leverages the origin of each WSI, such as lung, colon, or kidney, and uses GPT-4's internal knowledge to generate 20 potential microscopic image attributes of these origins, such as enlarged nuclei and lymphocyte infiltration. We calculate the similarity of all patches in the WSI for each prompt separately. For both report-based and attribute-based prompts, we identify the top 64 most relevant image patches in each WSI. Detailed prompts for GPT-4 can be found in supplementary materials.
\textbf{\textit{(2) K-means clustering:}} Since the prompt-based retrieval primarily focuses on patches with a higher degree of pathological changes, potentially missing a more diverse set of images. To address this, we use PathGen-CLIP-L$_{init}$ to extract features from WSI patches and apply k-means clustering on these features. Each cluster represents patches with distinct morphological characteristics.
The number of clusters is determined by the square root of the total number of patches, as larger WSI typically contains more distinct morphological features. We sample 256 patches from each WSI using clustering, ensuring uniform sampling within each cluster across the WSI to guarantee a more diverse selection of samples. Finally, we combine the results from prompt-based retrieval and clustering, resulting in a total of 384 representative patches sampled from each WSI.

\noindent\textbf{Step2: Similar Patch Filtering:} Since the representative samples may still contain highly similar images, this could pose challenges for subsequent training in CLIP contrastive learning. To address this, we employ PathGen-CLIP$_{init}$  to exclude highly similar patches. We set a similarity threshold of 0.88, patches exceeding this threshold are probabilistically filtered out, with higher similarities leading to higher filtering probabilities. We provide filtering pseudocode in supplementary materials.

\noindent\textbf{Step3: Description Generation:} 
We utilize the trained PathGen-LLaVA$_{desp}$ model with the prompt: "This is a histopathology image from {source}, describe this image in detail," where {source} indicates the origin of the WSI (e.g., lung, colon). This approach is used to generate detailed descriptions for all representative patches extracted from the WSIs, thereby creating initial image-description pairs.

\noindent\textbf{Step4: Description Revision:} 
Considering the potential for errors or hallucinations by LMMs during the description generation process, which may introduce inaccuracies descriptions, we employ the trained revision agent to identify and modify any parts of the descriptions that require additions, edits, or deletions, thereby ensuring higher accuracy and reducing noise in the final descriptions.

\noindent\textbf{Step5: Description Summarization:} 
Since the descriptions generated by PathGen-LLaVA$_{desp}$ are usually lengthy and often exceed the CLIP 77-token limit, we use a summarization agent to extract the key information from these descriptions, ensuring that no essential details are lost.

Through these steps, we generate a total of 1.6 million high-quality image-text pairs from 7,300 WSIs, sourced from 27 different tissue types. These diverse pathology image-text pairs are utilized for the contrastive learning pre-training of the CLIP model. This effort aims to develop a more robust pathology-specific CLIP model, which enhances support for downstream pathology tasks.

%% file: sec/4_experiment.tex
\section{Experiments}

In this section, we first introduce the training of PathGen-CLIP and PathGen-LLaVA$_{desp}$ using our generated data. Then, we conduct various downstream pathology tasks to validate PathGen-CLIP's performance and compare it with several state-of-the-art baseline models. These downstream tasks include zero-shot image classification, few-shot image classification, and whole slide image classification. Finally, we integrate CLIP into LLMs to demonstrate that based on our data and the robust PathGen-CLIP backbone, we can achieve superior pathology LMMs.

\subsection{Implement Details of Model Training Process}
\textbf{PathGen-LLaVA$_{desp}$:} 
PathGen-LLaVA$_{desp}$ adopts LLaVA's model structure and training approaches, divided into two stages. Initially, we align PathGen-CLIP-L$_{init}$ with Vicuna LLM using image-text pairs from PathGen$_{init}$, facilitated by a fully connected (FC) layer. Subsequently, we fine-tune both the FC layer and the Vicuna component using pre-generated detailed image descriptions. This process equips PathGen-LLaVA$_{desp}$ with the capability to generate image descriptions.

\textbf{PathGen-CLIP:} 
The training of PathGen-CLIP involves a two-stage process: first, using PathGen-1.6M data for pre-training, followed by fine-tuning with PathGen$_{init}$ data. This approach is chosen because the data generated using PathGen-LLaVA$_{desp}$ predominantly features extensive morphological descriptions. Additionally, the LLM component (Vicuna) has undergone specialized value alignment training, often recommending consultation with a professional pathologist for definitive diagnoses. Therefore, we utilize PathGen-1.6M for the first stage of training to help the model learn key morphological and structural features. Subsequently, PathGen$_{init}$  is employed in the second stage of training to incorporate multimodal knowledge that includes diagnostic information. In the supplementary materials, we conduct ablation studies to compare the effects of merging PathGen$_{init}$  and PathGen-1.6M for simultaneous training versus using PathGen-1.6M for the first stage of training followed by PathGen$_{init}$ in the second stage.

\begin{table}[t!]
	\caption{Comparison of different CLIP models on various pathology image classification datasets with accuracy (\%). The top performance is highlighted in \textbf{bold}, with the second-best \underline{underlined}.}
	\resizebox{\linewidth}{!}{
		\begin{tabular}{@{}ccccccccccc@{}}
			\toprule
			\textbf{Model}                    & \textbf{LC-Lung} & \textbf{LC-Colon} & \textbf{CRC100K} & \textbf{SkinCancer} & \textbf{Pcam} & \textbf{BACH} & \textbf{Osteo} & \textbf{WSSSLUAD} & \textbf{SICAPv2} & \textbf{Average} \\ \midrule
			OpenAI-CLIP              & 33.1    & 75.7     & 26.2    & 9.6        & 53.9 & 21.7 & 46.9  & 64.6     & 32.8    & 40.6 \\
			OpenAI-CLIP-L              & 70.4    & 81.1     & 40.3    & 19.4        & 55.5 & 34.3 & 53.9  & 81.2     & 25.4    & 51.3 \\
   	PLIP                     & 87.9      & 90.2       & 52.8      & 42.5         & 51.8   & 34.3   & 52.9    & 73.1     & 42.5      & 58.6 \\
   	PubmedCLIP              & 33.3    & 80.5     & 31.5    & 11.3        & 65.4 & 34.8 & 30.0  & 65.4     & 7.0    & 39.8 \\
      	PMC-CLIP              & 33.3    & 51.9     & 8.7    & 11.4        & 53.8 & 21.3 & 29.2  & 65.2     & 31.5    & 34.0 \\
			QuiltNet                & 80.0      & 91.0       & 49.5      & 46.4         & 58.7   & 43.8   & 53.8    & 70.5     & 37.3      & 58.9 \\
			PathCLIP        & 88.9    & 94.3   & 55.3    & 35.1        & 72.5 & 46.8   & 69.2   & \textbf{85.1}     & 48.3    & 66.2 \\
			BiomedCLIP               & 48.8      & 94.3       & 29.9      & 31.7         & 84.0   & 39.8   & 36.7    & 73.7      & 32.2     & 52.9 \\
			
		\rowcolor{aliceblue}	PathGen-CLIP        & \textbf{90.0}    & \underline{97.5}   & \underline{63.3}    & \underline{65.6}        & \textbf{89.2} & \underline{59.5}   & \underline{73.5}   & \underline{82.9}     & \underline{50.3}    & \underline{74.3} \\
			\rowcolor{aliceblue} PathGen-CLIP-L      & \underline{89.8}    & \textbf{99.3}   & \textbf{78.0}    & \textbf{70.6}        & \underline{88.2} & \textbf{71.5}   & \textbf{74.6}   & 82.2     & \textbf{63.5}    & \textbf{79.7} \\
			\bottomrule
	\end{tabular}}
 \vspace{-1em}

\label{tab:zero-shot}
\end{table}

\subsection{Zero-shot Image Classification}
CLIP-based models are trained on image-text pairs using contrastive learning, resulting in an intrinsic alignment between textual descriptions and visual content. This alignment enables effective zero-shot image classification, particularly useful in scenarios without annotations. To underscore the capabilities of the PathGen-CLIP series, we evaluate its zero-shot image classification performance on nine pathology classification datasets, including PatchCamelyon (Pcam)~\cite{veeling2018rotation}, CRC-100K~\cite{kather2018100}, SICAPv2~\cite{silva2020going}, BACH~\cite{aresta2019bach}, Osteo~\cite{arunachalam2019viable}, SkinCancer~\cite{kriegsmann2022deep}, WSSSLUAD~\cite{han2022wsss4luad}, LC-Lung, and LC-Colon~\cite{borkowski2019lung}. For each dataset, we design class-specific prompts, such as ``a H\&E image of {class}", and calculate the similarity between each class's text prompt and the image. The class prompt with the highest similarity score is assigned as the predicted label. We compare the performance of PathGen-CLIP with eight previous CLIP models, including OpenAI-CLIP, OpenAI-CLIP-L, PLIP, PMC-CLIP, PubMedCLIP, QuiltNet, PathCLIP, and BiomedCLIP. 

\textit{\textbf{Results: The PathGen-CLIP series significantly outperforms previous SOTA models in zero-shot classification tasks, with PathGen-CLIP-L emerging as a particularly advanced model.}} As demonstrated in Table \ref{tab:zero-shot}, PathGen-CLIP exceeds QuiltNet by 30.5\% on the Pcam dataset and by 19.2\% on the SkinCancer dataset. On average performance across all datasets, PathGen-CLIP also far surpasses the previously best model, PathCLIP, by 8.1\%. Moreover, our most robust variant, the PathGen-CLIP-L model, exhibits exceptionally consistent performance across various datasets, achieving remarkable results even in datasets where all other models perform poorly. For instance, on the BACH dataset, while PathGen-CLIP already surpasses previous models by a large margin (12.7\%), PathGen-CLIP-L further improves upon this by an additional 12\%. Similarly, it exceeds the average performance of PathGen-CLIP by 5.4\%. The high performance of both PathGen-CLIP and PathGen-CLIP-L underscores the effectiveness of our PathGen-1.6M dataset, offering potential for clinical utility in scenarios where no annotated data is available.

\begin{figure}        {\centerline{\includegraphics[width=\linewidth]{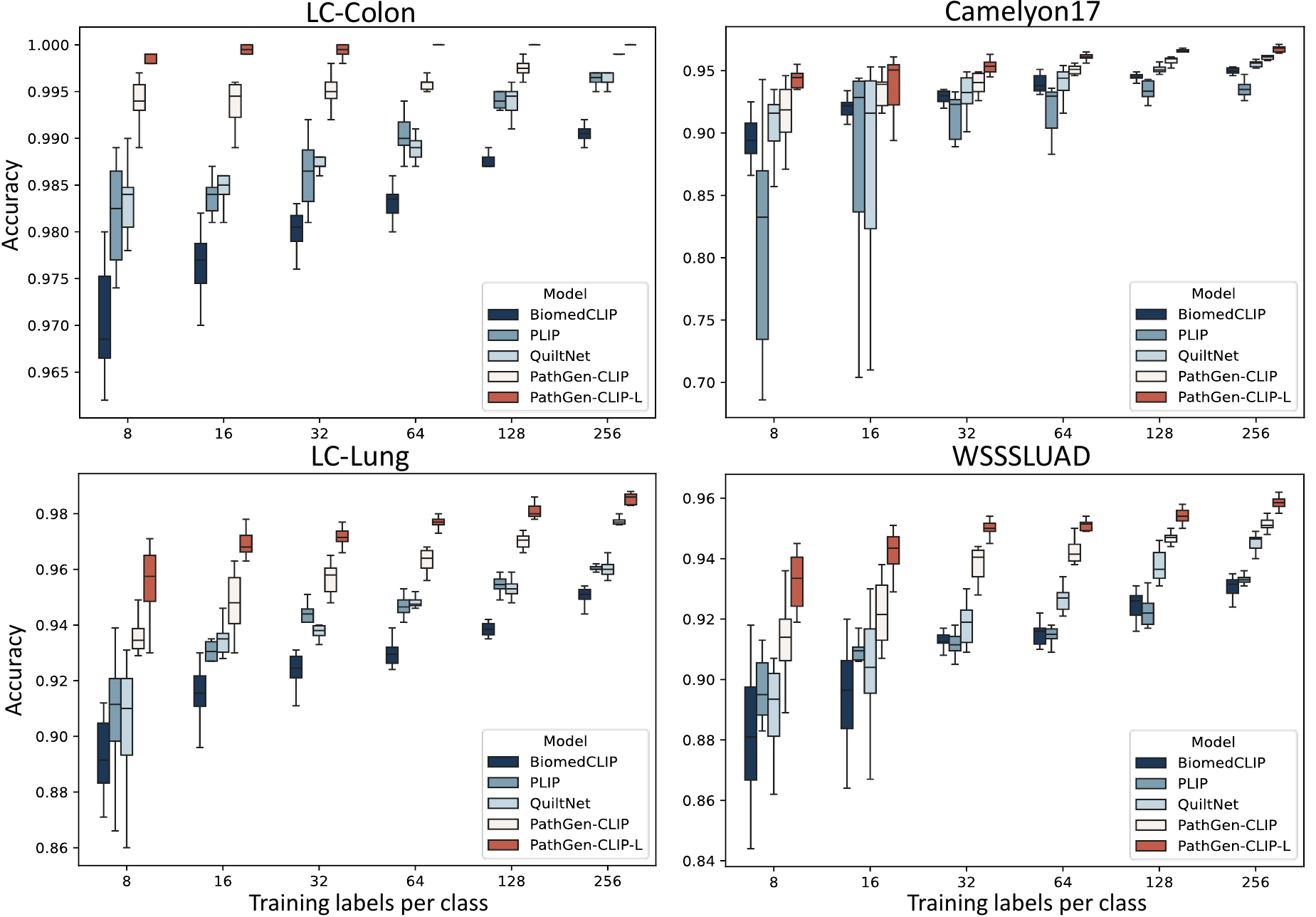}}}

	\caption{Comparison of few-shot classification using linear probing with different CLIP models on various pathology image classification datasets with accuracy (\%).}
	\label{fig:few_shot_classification}
\end{figure}

\subsection{Few-shot Image Classification with Linear Probing}
Traditional image classification tasks generally require extensive labeled data to achieve high accuracy, which is often impractical in real-world applications due to resource constraints, especially in the pathology domain. In this study, we explore the effectiveness of PathGen-CLIP in a few-shot setting, where it is fine-tuned using linear probing. This involves freezing the encoder and adding a FC layer on top, allowing us to evaluate the quality of the features generated by the model.
We assess the model on four representative datasets: LC-Colon, Camelyon17, LC-Lung, and WSSSLUAD, with fine-tuning performed using various training sizes (8, 16, 32, 64, 128, and 256 shots). Each training size is randomly sampled 10 times to train the model, avoiding unfairness due to randomness. The results are displayed in box plots to illustrate the model's performance fluctuations on the test set using different sampled training sets.

\textit{\textbf{Results: The PathGen-CLIP series quickly achieves remarkable performance with minimal samples, making it viable for clinical settings with limited annotations.}} As illustrated in Figure \ref{fig:few_shot_classification}, with only 8 shots, PathGen-CLIP reaches nearly 94\% accuracy on the LC-Lung dataset, significantly outperforming competitors like PLIP, QuiltNet, and BiomedCLIP, which achieve accuracies below 92\%. In this scenario, the PathGen-CLIP-L version even reaches 96\% accuracy. Furthermore, as the number of shots increases, our model consistently outperforms other baseline models. Additionally, the narrower box plot widths of the PathGen-CLIP model illustrate its superior task adaptation capabilities and robustness.

\begin{table*}[t]
	\centering
	\caption{Performance of various CLIP models across three datasets, two MIL methods, and two evaluation metrics. The top performance is highlighted in \textbf{bold}, with the second-best \underline{underlined}.}
	\resizebox{\linewidth}{!}{
		\begin{tabular}{clcccccccc}
			\toprule
			\multicolumn{2}{c}{\multirow{2}{*}{\diagbox[width=10em]{{Method}}{{Performance}}}} & \multicolumn{2}{c}{\textbf{CAMELYON-17}} & \multicolumn{2}{c}{\textbf{CAMELYON-16}} &  \multicolumn{2}{c}{\textbf{BRACS}} & \multicolumn{2}{c}{\textbf{Average}} \\ 
			\cmidrule(lr){3-4} \cmidrule(lr){5-6} \cmidrule(lr){7-8} \cmidrule(lr){9-10}
			& & F1-score & AUC & F1-score & AUC & F1-score & AUC & F1-score & AUC \\
			\midrule
			\multirow{7}{*}{\rotatebox{90}{\makecell{ABMIL}}} &
			OpenAI-CLIP & 23.5\tiny{$\pm$4.6} & 60.7\tiny{$\pm$3.3} & 62.8\tiny{$\pm$3.5} & 61.4\tiny{$\pm$2.7} & 46.8\tiny{$\pm$0.5} & 78.5\tiny{$\pm$0.7} & 44.4 & 66.9 \\
			& OpenAI-CLIP-L & 37.7\tiny{$\pm$2.9} & 76.7\tiny{$\pm$2.1} & 75.8\tiny{$\pm$0.8} & 69.7\tiny{$\pm$1.6} & 51.6\tiny{$\pm$3.9} & 78.9\tiny{$\pm$0.9} & 55.0 & 75.1 \\
			& PLIP & 45.6\tiny{$\pm$5.0} & 82.8\tiny{$\pm$1.1} & 86.6\tiny{$\pm$1.4} & 90.0\tiny{$\pm$2.7} & 51.7\tiny{$\pm$2.0} & 78.5\tiny{$\pm$0.4} & 61.3 & 83.8 \\
			& Quilt-Net & 44.3\tiny{$\pm$2.0} & 84.2\tiny{$\pm$1.0} & 82.9\tiny{$\pm$1.6} & 87.1\tiny{$\pm$2.2} & 54.7\tiny{$\pm$3.0} & 82.3\tiny{$\pm$1.6} & 60.6 & 84.5 \\
			& BiomedCLIP & 55.5\tiny{$\pm$2.5} & 84.1\tiny{$\pm$1.2} & 82.6\tiny{$\pm$1.4} & 83.6\tiny{$\pm$3.4} & 57.4\tiny{$\pm$4.8} & 80.6\tiny{$\pm$1.4} & 65.2 & 82.8 \\
			& PathCLIP & 45.2\tiny{$\pm$3.3} & 82.6\tiny{$\pm$1.9} & 80.2\tiny{$\pm$1.5} & 85.0\tiny{$\pm$1.5} & 56.4\tiny{$\pm$3.2} & 83.8\tiny{$\pm$0.8} & 60.6 & 83.8 \\
			\rowcolor{aliceblue}& PathGen-CLIP (ours) & \underline{58.2\tiny{$\pm$3.3}} & \underline{87.5\tiny{$\pm$1.1}} & \underline{93.5\tiny{$\pm$2.2}} & \underline{\textbf{96.9\tiny{$\pm$1.5}}} & \underline{62.6\tiny{$\pm$1.9}} & \underline{85.8\tiny{$\pm$0.8}} & \underline{71.4} & \underline{90.1} \\
			\rowcolor{aliceblue} & PathGen-CLIP-L (ours) & \textbf{58.6\tiny{$\pm$6.7}} & \textbf{87.9\tiny{$\pm$1.2}} & \textbf{94.3\tiny{$\pm$1.8}} & 95.8\tiny{$\pm$1.4} & \textbf{66.6\tiny{$\pm$6.4}} & \textbf{87.2\tiny{$\pm$2.9}} & \textbf{73.2} & \textbf{90.3} \\
			\midrule
			\multirow{7}{*}{\rotatebox{90}{\makecell{ACMIL}}} &
			OpenAI-CLIP & 25.4\tiny{$\pm$4.1} & 59.4\tiny{$\pm$3.7} & 63.7\tiny{$\pm$4.3} & 67.3\tiny{$\pm$4.3} & 52.2\tiny{$\pm$4.5} & 76.7\tiny{$\pm$1.6} & 47.1 & 67.8 \\
			& OpenAI-CLIP-L & 34.9\tiny{$\pm$4.5} & 78.5\tiny{$\pm$3.0} & 78.7\tiny{$\pm$3.7} & 73.4\tiny{$\pm$3.4} & 55.5\tiny{$\pm$3.6} & 78.8\tiny{$\pm$2.0} & 56.4 & 76.9 \\
			& PLIP & 46.0\tiny{$\pm$1.5} & 86.1\tiny{$\pm$1.0} & 90.4\tiny{$\pm$2.2} & 94.8\tiny{$\pm$1.3} & 57.0\tiny{$\pm$3.3} & 80.8\tiny{$\pm$0.7} & 64.5 & 87.2 \\
			& Quilt-Net & 44.4\tiny{$\pm$1.0} & 86.0\tiny{$\pm$0.9} & 84.8\tiny{$\pm$3.5} & 90.5\tiny{$\pm$3.4} & 60.8\tiny{$\pm$3.5} & 82.0\tiny{$\pm$1.9} & 63.3 & 86.2 \\
			& BiomedCLIP & \underline{53.6\tiny{$\pm$4.0}} & 83.6\tiny{$\pm$1.3} & 82.9\tiny{$\pm$2.0} & 84.8\tiny{$\pm$3.3} & 63.2\tiny{$\pm$2.0} & 81.9\tiny{$\pm$0.8} & 66.6 & 83.4 \\
			& PathCLIP & 44.2\tiny{$\pm$0.6} & 82.9\tiny{$\pm$1.3} & 84.4\tiny{$\pm$1.6} & 87.3\tiny{$\pm$1.6} & 58.0\tiny{$\pm$7.6} & 83.0\tiny{$\pm$2.4} & 62.2 & 84.4 \\
			\rowcolor{aliceblue}& PathGen-CLIP (ours) & 53.3\tiny{$\pm$4.6} & \underline{89.4\tiny{$\pm$1.2}} & \underline{92.6\tiny{$\pm$1.6}} & \underline{97.2\tiny{$\pm$0.9}} & \textbf{66.9\tiny{$\pm$3.0}} & \underline{87.0\tiny{$\pm$0.4}} & \underline{71.0} & \underline{91.2} \\
			\rowcolor{aliceblue} & PathGen-CLIP-L (ours) & \textbf{58.4\tiny{$\pm$5.2}} & \textbf{92.0\tiny{$\pm$0.7}} & \textbf{94.5\tiny{$\pm$1.0}} & \textbf{97.4\tiny{$\pm$1.9}} & \textbf{66.9\tiny{$\pm$5.0}} & \textbf{88.4\tiny{$\pm$1.4}} & \textbf{73.3} & \textbf{92.6} \\
			\bottomrule
		\end{tabular}
	}
	\label{tab:WSI_classification}
\end{table*}

\begin{table*}[h]
	\caption{Overall results of models on the PathMMU \textbf{test set}. The best-performing LMM in each subset for general and pathology domain LMMs is \textbf{in-bold}, and the top-performing LMM is {\underline{underlined}}.}
	\centering
	\resizebox{\linewidth}{!}{
		\begin{tabular}{@{}lcccccccccccc@{}}
			\toprule
			\textbf{} & \multicolumn{2}{c}{\textbf{Test Overall}} & \multicolumn{2}{c}{\textbf{PubMed}} & \multicolumn{2}{c}{\textbf{SocialPath}} & \multicolumn{2}{c}{\textbf{EduContent}} & \multicolumn{2}{c}{\textbf{Atlas}} &\multicolumn{2}{c}{\textbf{PathCLS}} \\ 
			& Tiny  & ALL  & Tiny  & ALL & Tiny  & All & Tiny  & All  & Tiny  & ALL  & Tiny  & ALL\\
			& (1156)  & (9677)  & (281)  & (3068) & (235)  & (1855) & (255)  & (1938)  & (208)  & (1007) & (177)  & (1809)\\\midrule
	\color{gray} Random Choice &  \color{gray} 22.1 & \color{gray} 23.7 & \color{gray} 22.1 & \color{gray} 25.1 & \color{gray} 25.5 & \color{gray} 26.5 & \color{gray} 25.5 & \color{gray} 26.0 & \color{gray} 19.7 & \color{gray} 23.0  & \color{gray} 15.3 & \color{gray} 16.3\\ 
			\color{gray} Frequent Choice &  \color{gray} 27.7 & \color{gray} 25.5 & \color{gray} 28.8 & \color{gray} 26.1 & \color{gray} 27.7 & \color{gray} 26.7 & \color{gray} 29.8 & \color{gray} 26.5 & \color{gray} 28.4 & \color{gray} 27.5 & \color{gray} 22.0 & \color{gray} 21.0  \\
			\rowcolor{dino}  Expert performance &  71.8 &  - &  72.9 & -& 71.5 &  - &  69.0 &  - & 68.3 &  - & 78.9 &  - \\
			\midrule
			\multicolumn{13}{c}{\textbf{General Large Multimodal Models}} \\ \midrule
		BLIP-2 FLAN-T5-XXL       & 33.3 & 33.5 & 37.0 & 37.4 & 35.7 & 34.6 & 30.2 & 34.5 & 39.4 & 40.7 & 19.8 & 20.6  \\
		InstructBLIP-FLAN-T5-XXL & 34.3 & 33.9 & 39.1 & 37.2 & 33.6 & 34.3 & 34.5 & 36.0 & 38.5 & 39.3 & 22.6 & 22.7  \\
			LLaVA-1.5-13B            & 38.8 & 37.6 & 44.5 & 41.0 & 40.4 & 40.4 & 34.1 & 39.4 & 47.1 & 44.3 & 24.9 & 23.5  \\
			Qwen-VL-PLUS       & 39.3 & 34.3 & 43.5 & 37.7 & 41.3 & 36.0 & 39.6 & 36.0 & 44.7 & 37.1 & 23.2 & 23.3  \\
			Qwen-VL-MAX        & 49.2 & 45.9 & 53.0 & 50.9 & 53.6 & 49.3 & 52.2 & 47.9  & \underline{51.4} & 49.8 & 30.5 & 29.6 \\
			Gemini Pro Vision  & 42.8 & 42.7 & 43.8 & 44.9 & 42.4 & 42.0 & 43.5 & 43.7 & 49.5 & 49.4 & 32.8 & \underline{34.7}   \\
			GPT-4V-1106             & \underline{53.9} & \underline{49.8} & \underline{59.4} & \underline{53.5} & \underline{58.7} & \underline{53.9} & \underline{60.4} & \underline{53.6} & 48.1 & \underline{52.8}  & \underline{36.2} & 33.8\\
			\midrule
			\multicolumn{13}{c}{\textbf{Pathology-specific Large Multimodal Models}} \\ \midrule
			LLaVA-Med & 25.3 & 26.2 & 28.5 & 27.7 & 28.9 & 27.3 & 22.7 & 27.2 & 22.6 & 30.7 & 22.6 & 20.3\\ 			
			Quilt-LLaVA    & 45.6 & 41.5 & 47.3 & 42.6 & 46.4 & 46.6 & 51.8 & 45.3 & 46.2 & 42.7 & 32.2 & 29.2\\
			\rowcolor{aliceblue} PathGen-LLaVA & \textbf{60.1} & \textbf{58.4} & \textbf{60.1} & \textbf{60.1} & \textbf{60.9} & \textbf{58.8} & \textbf{60.8} & \textbf{60.7} & \textbf{63.5} & \textbf{64.9} & \textbf{54.2} & \textbf{48.9}\\     
			\bottomrule
	\end{tabular}}
	\label{tab:overall_results_pathmmu}
\end{table*}

\subsection{Whole Slide Image Classification}

Whole slide image classification is essential for automating disease identification and classification from high-resolution pathological slides by analyzing high-resolution pathological slide images, which are typically larger than 10,000 $\times$ 10,000 pixels. This task is particularly valuable for clinical practitioners as it boosts clinical accuracy and efficiency significantly. The standard approach for WSI classification involves segmenting WSIs into image patches, extracting instance embeddings using a frozen image encoder, and employing Multiple Instance Learning (MIL) to convert these embeddings into slide-level predictions. Superior patch representations provided by the image encoder are crucial, as they significantly influence WSI prediction performance. Therefore, we assess the efficacy of the PathGen-CLIP series in comparison with other prominent models, including OpenAI-CLIP, OpenAI-CLIP-L, PLIP, BiomedCLIP, PathCLIP, and Quilt-Net. For the MIL method, we utilize the widely adopted ABMIL \cite{ilse2018attention} and the current SOTA method, ACMIL \cite{zhang2023attention}. Our evaluations span three datasets: CAMELYON16 \cite{litjens20181399}, CAMELYON17 \cite{litjens20181399}, and BRACS \cite{brancati2022bracs}, excluding any TCGA-related datasets. For the detailed experimental setup, please refer to the supplementary materials.

\textbf{\textit{Results: PathGen-CLIP series consistently and significantly outperform existing pathology-specific CLIP models across three key datasets leveraging both MIL methodologies (ABMIL and ACMIL).}}
For instance, as shown in Table \ref{tab:WSI_classification}, the PathGen-CLIP model, utilizing the ABMIL architecture, achieves an impressive average AUC of 96.9 on the CAMELYON16 dataset. This performance significantly surpasses that of PLIP at 90.0, BiomedCLIP at 83.6, and Quilt-Net at 87.1. Overall, PathGen-CLIP models demonstrate considerable enhancements; the average AUC across datasets when employing the ACMIL architecture is 91.2, exceeding the performances of PLIP at 87.2, BiomedCLIP at 83.4, and Quilt-Net at 86.2. The larger variant, PathGen-CLIP-L, achieves an even more remarkable average AUC of 92.6, highlighting the substantial advancements our models bring to the analysis of whole slide images.

\subsection{Integrating with Large Language Models}

LLMs possess extensive knowledge and common sense due to their substantial model sizes and comprehensive training datasets. Consequently, models like CLIP, already aligned with language models, are frequently used for integration with LLMs. This integration leverages the broader and powerful capabilities of LLMs to construct versatile LMMs. In this work, to train such LMMs, we construct 200K instruction-tuning samples based on PathGen data, including 95K pathology multimodal multi-choice QAs and 105K rounds of dialogue data. We follow the training methodology of LLaVA to train our pathology LMM, which we refer to as PathGen-LLaVA. We assess the performance of PathGen-LLaVA using the PathMMU dataset, which includes expert annotations from various sources and diseases. Additionally, we benchmark PathGen-LLaVA against the most advanced general-domain models, including GPT-4V, Gemini-Pro Vision, Qwen-VL-Max, as well as previous pathology-specific LMMs such as LLaVA-Med and Quilt-LLaVA, to validate the capabilities of our model. For detailed information on the construction of training data and model training details, please refer to the supplementary materials.

\textit{\textbf{Results: PathGen-LLaVA significantly outperforms previous SOTA pathology LMMs, even surpassing the leading general model, GPT-4V.}} As shown in Table \ref{tab:overall_results_pathmmu}, PathGen-LLaVA consistently exceeds Quilt-LLaVA by 17.5\%, 12.2\%, 15.4\%, and 22.2\% across the PathMMU's PubMed, SocialPath, Atlas, and PathCLS subsets, respectively. Moreover, it also significantly outperforms the top general model, GPT-4V, by a notable margin. For instance, in overall test performance, it achieved 58.4\%, surpassing GPT-4V’s 49.8\%. This underscores the superiority of the PathGen dataset and the effectiveness of PathGen-CLIP as a vision backbone for integration with LLMs. It marks a step closer to the potential application of pathology LMMs in assisting medical professionals in practice.

%% file: sec/5_conclusion.tex
\section{Conclusion and Limitations}
\label{sec:conclu_limit}
In this work, we propose a novel approach leveraging multi-agent collaboration to generate 1.6 million high-quality pathology image-text pairs from WSIs. Utilizing these generated data and existing datasets, we train two more powerful models: PathGen-CLIP and PathGen-CLIP-L. These models achieve significant advancements in zero-shot image classification, few-shot image classification, and whole slide image classification. Notably, PathGen-CLIP-L is the first large version of the CLIP model introduced in the pathology domain, markedly enhancing capabilities for various tasks. Additionally, we integrate PathGen-CLIP-L with an LLM to create a more robust pathology LMM, named PathGen-LLaVA. By leveraging the PathGen dataset, we generate 200,000 instruction tuning samples to train PathGen-LLaVA. Our experiments demonstrate that PathGen-LLaVA exhibits robust pathology image understanding capabilities, significantly outperforming previous pathology LMMs on the large-scale PathMMU dataset by a large margin and even surpassing the performance of GPT-4V. These extensive experiments confirm the superiority and potential of the PathGen dataset, and we are optimistic about its ability to bring significant advancements to the field of pathology. However, a limitation is that our approach partially relies on WSI reports, which are difficult to obtain. Therefore, future work should investigate whether satisfactory results can be achieved without relying on WSI reports and further explore the potential impacts of data expansion on the model.

%% file: neurips_data_2024.bbl
\begin{thebibliography}{10}

\bibitem{alayrac2022flamingo}
Jean-Baptiste Alayrac, Jeff Donahue, Pauline Luc, Antoine Miech, Iain Barr, Yana Hasson, Karel Lenc, Arthur Mensch, Katherine Millican, Malcolm Reynolds, et~al.
\newblock Flamingo: a visual language model for few-shot learning.
\newblock In {\em Advances in Neural Information Processing Systems}, 2022.

\bibitem{aresta2019bach}
Guilherme Aresta, Teresa Ara{\'u}jo, Scotty Kwok, Sai~Saketh Chennamsetty, Mohammed Safwan, Varghese Alex, Bahram Marami, Marcel Prastawa, Monica Chan, Michael Donovan, et~al.
\newblock Bach: Grand challenge on breast cancer histology images.
\newblock {\em Medical image analysis}, 56:122--139, 2019.

\bibitem{arunachalam2019viable}
Harish~Babu Arunachalam, Rashika Mishra, Ovidiu Daescu, Kevin Cederberg, Dinesh Rakheja, Anita Sengupta, David Leonard, Rami Hallac, and Patrick Leavey.
\newblock Viable and necrotic tumor assessment from whole slide images of osteosarcoma using machine-learning and deep-learning models.
\newblock {\em PloS one}, 14(4):e0210706, 2019.

\bibitem{bai2023qwen}
Jinze Bai, Shuai Bai, Shusheng Yang, Shijie Wang, Sinan Tan, Peng Wang, Junyang Lin, Chang Zhou, and Jingren Zhou.
\newblock Qwen-vl: A frontier large vision-language model with versatile abilities.
\newblock {\em arXiv preprint arXiv:2308.12966}, 2023.

\bibitem{fuyu-8b}
Rohan Bavishi, Erich Elsen, Curtis Hawthorne, Maxwell Nye, Augustus Odena, Arushi Somani, and Sa\u{g}nak Ta\c{s}\i{}rlar.
\newblock Introducing our multimodal models, 2023.

\bibitem{borkowski2019lung}
Andrew~A Borkowski, Marilyn~M Bui, L~Brannon Thomas, Catherine~P Wilson, Lauren~A DeLand, and Stephen~M Mastorides.
\newblock Lung and colon cancer histopathological image dataset (lc25000).
\newblock {\em arXiv preprint arXiv:1912.12142}, 2019.

\bibitem{brancati2022bracs}
Nadia Brancati, Anna~Maria Anniciello, Pushpak Pati, Daniel Riccio, Giosu{\`e} Scognamiglio, Guillaume Jaume, Giuseppe De~Pietro, Maurizio Di~Bonito, Antonio Foncubierta, Gerardo Botti, et~al.
\newblock Bracs: A dataset for breast carcinoma subtyping in h\&e histology images.
\newblock {\em Database}, 2022:baac093, 2022.

\bibitem{instructblip}
Wenliang Dai, Junnan Li, Dongxu Li, Anthony Meng~Huat Tiong, Junqi Zhao, Weisheng Wang, Boyang Li, Pascale Fung, and Steven Hoi.
\newblock Instructblip: Towards general-purpose vision-language models with instruction tuning, 2023.

\bibitem{dai2024instructblip}
Wenliang Dai, Junnan Li, Dongxu Li, Anthony Meng~Huat Tiong, Junqi Zhao, Weisheng Wang, Boyang Li, Pascale~N Fung, and Steven Hoi.
\newblock Instructblip: Towards general-purpose vision-language models with instruction tuning.
\newblock {\em Advances in Neural Information Processing Systems}, 36, 2024.

\bibitem{eslami2023pubmedclip}
Sedigheh Eslami, Christoph Meinel, and Gerard De~Melo.
\newblock Pubmedclip: How much does clip benefit visual question answering in the medical domain?
\newblock In {\em Findings of the Association for Computational Linguistics: EACL 2023}, pages 1181--1193, 2023.

\bibitem{esmaeilpour2022zero}
Sepideh Esmaeilpour, Bing Liu, Eric Robertson, and Lei Shu.
\newblock Zero-shot out-of-distribution detection based on the pre-trained model clip.
\newblock In {\em Proceedings of the AAAI conference on artificial intelligence}, volume~36, pages 6568--6576, 2022.

\bibitem{guo2024histgen}
Zhengrui Guo, Jiabo Ma, Yingxue Xu, Yihui Wang, Liansheng Wang, and Hao Chen.
\newblock Histgen: Histopathology report generation via local-global feature encoding and cross-modal context interaction.
\newblock {\em arXiv preprint arXiv:2403.05396}, 2024.

\bibitem{han2022wsss4luad}
Chu Han, Xipeng Pan, Lixu Yan, Huan Lin, Bingbing Li, Su~Yao, Shanshan Lv, Zhenwei Shi, Jinhai Mai, Jiatai Lin, et~al.
\newblock Wsss4luad: Grand challenge on weakly-supervised tissue semantic segmentation for lung adenocarcinoma.
\newblock {\em arXiv preprint arXiv:2204.06455}, 2022.

\bibitem{hong2023metagpt}
Sirui Hong, Xiawu Zheng, Jonathan Chen, Yuheng Cheng, Ceyao Zhang, Zili Wang, Steven Ka~Shing Yau, Zijuan Lin, Liyang Zhou, Chenyu Ran, et~al.
\newblock Metagpt: Meta programming for multi-agent collaborative framework.
\newblock {\em arXiv preprint arXiv:2308.00352}, 2023.

\bibitem{hua2023war}
Wenyue Hua, Lizhou Fan, Lingyao Li, Kai Mei, Jianchao Ji, Yingqiang Ge, Libby Hemphill, and Yongfeng Zhang.
\newblock War and peace (waragent): Large language model-based multi-agent simulation of world wars, 2023.

\bibitem{huang2023visual}
Zhi Huang, Federico Bianchi, Mert Yuksekgonul, Thomas~J Montine, and James Zou.
\newblock A visual--language foundation model for pathology image analysis using medical twitter.
\newblock {\em Nature medicine}, 29(9):2307--2316, 2023.

\bibitem{ikezogwo2024quilt}
Wisdom Ikezogwo, Saygin Seyfioglu, Fatemeh Ghezloo, Dylan Geva, Fatwir Sheikh~Mohammed, Pavan~Kumar Anand, Ranjay Krishna, and Linda Shapiro.
\newblock Quilt-1m: One million image-text pairs for histopathology.
\newblock {\em Advances in Neural Information Processing Systems}, 36, 2024.

\bibitem{ilse2018attention}
Maximilian Ilse, Jakub Tomczak, and Max Welling.
\newblock Attention-based deep multiple instance learning.
\newblock In {\em International conference on machine learning}, pages 2127--2136. PMLR, 2018.

\bibitem{jia2021scaling}
Chao Jia, Yinfei Yang, Ye~Xia, Yi-Ting Chen, Zarana Parekh, Hieu Pham, Quoc Le, Yun-Hsuan Sung, Zhen Li, and Tom Duerig.
\newblock Scaling up visual and vision-language representation learning with noisy text supervision.
\newblock In {\em International conference on machine learning}, pages 4904--4916. PMLR, 2021.

\bibitem{kather2018100}
Jakob~Nikolas Kather, Niels Halama, and Alexander Marx.
\newblock 100,000 histological images of human colorectal cancer and healthy tissue.
\newblock {\em Zenodo10}, 5281, 2018.

\bibitem{kriegsmann2022deep}
Katharina Kriegsmann, Frithjof Lobers, Christiane Zgorzelski, Joerg Kriegsmann, Charlotte Janssen, Rolf~Ruedinger Meliss, Thomas Muley, Ulrich Sack, Georg Steinbuss, and Mark Kriegsmann.
\newblock Deep learning for the detection of anatomical tissue structures and neoplasms of the skin on scanned histopathological tissue sections.
\newblock {\em Frontiers in Oncology}, 12:1022967, 2022.

\bibitem{kumar2014robbins}
Vinay Kumar, Abul~K Abbas, Nelson Fausto, and Jon~C Aster.
\newblock {\em Robbins and Cotran pathologic basis of disease, professional edition e-book}.
\newblock Elsevier health sciences, 2014.

\bibitem{li2023llava}
Chunyuan Li, Cliff Wong, Sheng Zhang, Naoto Usuyama, Haotian Liu, Jianwei Yang, Tristan Naumann, Hoifung Poon, and Jianfeng Gao.
\newblock Llava-med: Training a large language-and-vision assistant for biomedicine in one day.
\newblock {\em arXiv preprint arXiv:2306.00890}, 2023.

\bibitem{li2023blip}
Junnan Li, Dongxu Li, Silvio Savarese, and Steven Hoi.
\newblock Blip-2: Bootstrapping language-image pre-training with frozen image encoders and large language models.
\newblock In {\em International conference on machine learning}, pages 19730--19742. PMLR, 2023.

\bibitem{li2022blip}
Junnan Li, Dongxu Li, Caiming Xiong, and Steven Hoi.
\newblock Blip: Bootstrapping language-image pre-training for unified vision-language understanding and generation.
\newblock In {\em International conference on machine learning}, pages 12888--12900. PMLR, 2022.

\bibitem{lin2023pmc}
Weixiong Lin, Ziheng Zhao, Xiaoman Zhang, Chaoyi Wu, Ya~Zhang, Yanfeng Wang, and Weidi Xie.
\newblock Pmc-clip: Contrastive language-image pre-training using biomedical documents.
\newblock In {\em International Conference on Medical Image Computing and Computer-Assisted Intervention}, pages 525--536. Springer, 2023.

\bibitem{lin2023clip}
Yuqi Lin, Minghao Chen, Wenxiao Wang, Boxi Wu, Ke~Li, Binbin Lin, Haifeng Liu, and Xiaofei He.
\newblock Clip is also an efficient segmenter: A text-driven approach for weakly supervised semantic segmentation.
\newblock In {\em Proceedings of the IEEE/CVF Conference on Computer Vision and Pattern Recognition}, pages 15305--15314, 2023.

\bibitem{litjens20181399}
Geert Litjens, Peter Bandi, Babak Ehteshami~Bejnordi, Oscar Geessink, Maschenka Balkenhol, Peter Bult, Altuna Halilovic, Meyke Hermsen, Rob Van~de Loo, Rob Vogels, et~al.
\newblock 1399 h\&e-stained sentinel lymph node sections of breast cancer patients: the camelyon dataset.
\newblock {\em GigaScience}, 7(6):giy065, 2018.

\bibitem{liu2023improved}
Haotian Liu, Chunyuan Li, Yuheng Li, and Yong~Jae Lee.
\newblock Improved baselines with visual instruction tuning.
\newblock {\em arXiv preprint arXiv:2310.03744}, 2023.

\bibitem{liu2024visual}
Haotian Liu, Chunyuan Li, Qingyang Wu, and Yong~Jae Lee.
\newblock Visual instruction tuning.
\newblock {\em Advances in neural information processing systems}, 36, 2024.

\bibitem{lu2024visual}
Ming~Y Lu, Bowen Chen, Drew~FK Williamson, Richard~J Chen, Ivy Liang, Tong Ding, Guillaume Jaume, Igor Odintsov, Long~Phi Le, Georg Gerber, et~al.
\newblock A visual-language foundation model for computational pathology.
\newblock {\em Nature Medicine}, 30(3):863--874, 2024.

\bibitem{gpt4}
OpenAI.
\newblock Gpt-4 technical report, 2023.

\bibitem{openai2023gpt4v}
OpenAI.
\newblock Gpt-4v(ision) system card.
\newblock \url{https://cdn.openai.com/papers/GPTV_System_Card.pdf}, 2023.

\bibitem{park2023generative}
Joon~Sung Park, Joseph~C O'Brien, Carrie~J Cai, Meredith~Ringel Morris, Percy Liang, and Michael~S Bernstein.
\newblock Generative agents: Interactive simulacra of human behavior.
\newblock {\em arXiv preprint arXiv:2304.03442}, 2023.

\bibitem{park2022social}
Joon~Sung Park, Lindsay Popowski, Carrie Cai, Meredith~Ringel Morris, Percy Liang, and Michael~S Bernstein.
\newblock Social simulacra: Creating populated prototypes for social computing systems.
\newblock In {\em Proceedings of the 35th Annual ACM Symposium on User Interface Software and Technology}, pages 1--18, 2022.

\bibitem{qian2023communicative}
Chen Qian, Xin Cong, Wei Liu, Cheng Yang, Weize Chen, Yusheng Su, Yufan Dang, Jiahao Li, Juyuan Xu, Dahai Li, Zhiyuan Liu, and Maosong Sun.
\newblock Communicative agents for software development, 2023.

\bibitem{radford2021learning}
Alec Radford, Jong~Wook Kim, Chris Hallacy, Aditya Ramesh, Gabriel Goh, Sandhini Agarwal, Girish Sastry, Amanda Askell, Pamela Mishkin, Jack Clark, et~al.
\newblock Learning transferable visual models from natural language supervision.
\newblock In {\em International conference on machine learning}, pages 8748--8763. PMLR, 2021.

\bibitem{raffel2020exploring}
Colin Raffel, Noam Shazeer, Adam Roberts, Katherine Lee, Sharan Narang, Michael Matena, Yanqi Zhou, Wei Li, and Peter~J Liu.
\newblock Exploring the limits of transfer learning with a unified text-to-text transformer.
\newblock {\em The Journal of Machine Learning Research}, 21(1):5485--5551, 2020.

\bibitem{rombach2022high}
Robin Rombach, Andreas Blattmann, Dominik Lorenz, Patrick Esser, and Bj{\"o}rn Ommer.
\newblock High-resolution image synthesis with latent diffusion models.
\newblock In {\em Proceedings of the IEEE/CVF conference on computer vision and pattern recognition}, pages 10684--10695, 2022.

\bibitem{schuhmann2022laion}
Christoph Schuhmann, Romain Beaumont, Richard Vencu, Cade Gordon, Ross Wightman, Mehdi Cherti, Theo Coombes, Aarush Katta, Clayton Mullis, Mitchell Wortsman, et~al.
\newblock Laion-5b: An open large-scale dataset for training next generation image-text models.
\newblock {\em Advances in Neural Information Processing Systems}, 35:25278--25294, 2022.

\bibitem{seyfioglu2023quilt}
Mehmet~Saygin Seyfioglu, Wisdom~O Ikezogwo, Fatemeh Ghezloo, Ranjay Krishna, and Linda Shapiro.
\newblock Quilt-llava: Visual instruction tuning by extracting localized narratives from open-source histopathology videos.
\newblock {\em arXiv preprint arXiv:2312.04746}, 2023.

\bibitem{shu2023clipood}
Yang Shu, Xingzhuo Guo, Jialong Wu, Ximei Wang, Jianmin Wang, and Mingsheng Long.
\newblock Clipood: Generalizing clip to out-of-distributions.
\newblock In {\em International Conference on Machine Learning}, pages 31716--31731. PMLR, 2023.

\bibitem{silva2020going}
Julio Silva-Rodr{\'\i}guez, Adri{\'a}n Colomer, Mar{\'\i}a~A Sales, Rafael Molina, and Valery Naranjo.
\newblock Going deeper through the gleason scoring scale: An automatic end-to-end system for histology prostate grading and cribriform pattern detection.
\newblock {\em Computer methods and programs in biomedicine}, 195:105637, 2020.

\bibitem{sun2024pathmmu}
Yuxuan Sun, Hao Wu, Chenglu Zhu, Sunyi Zheng, Qizi Chen, Kai Zhang, Yunlong Zhang, Xiaoxiao Lan, Mengyue Zheng, Jingxiong Li, et~al.
\newblock Pathmmu: A massive multimodal expert-level benchmark for understanding and reasoning in pathology.
\newblock {\em arXiv preprint arXiv:2401.16355}, 2024.

\bibitem{sun2024pathasst}
Yuxuan Sun, Chenglu Zhu, Sunyi Zheng, Kai Zhang, Lin Sun, Zhongyi Shui, Yunlong Zhang, Honglin Li, and Lin Yang.
\newblock Pathasst: A generative foundation ai assistant towards artificial general intelligence of pathology.
\newblock In {\em Proceedings of the AAAI Conference on Artificial Intelligence}, volume~38, pages 5034--5042, 2024.

\bibitem{team2023gemini}
Gemini Team, Rohan Anil, Sebastian Borgeaud, Yonghui Wu, Jean-Baptiste Alayrac, Jiahui Yu, Radu Soricut, Johan Schalkwyk, Andrew~M Dai, Anja Hauth, et~al.
\newblock Gemini: a family of highly capable multimodal models.
\newblock {\em arXiv preprint arXiv:2312.11805}, 2023.

\bibitem{thomee2016yfcc100m}
Bart Thomee, David~A Shamma, Gerald Friedland, Benjamin Elizalde, Karl Ni, Douglas Poland, Damian Borth, and Li-Jia Li.
\newblock Yfcc100m: The new data in multimedia research.
\newblock {\em Communications of the ACM}, 59(2):64--73, 2016.

\bibitem{veeling2018rotation}
Bastiaan~S Veeling, Jasper Linmans, Jim Winkens, Taco Cohen, and Max Welling.
\newblock Rotation equivariant cnns for digital pathology.
\newblock In {\em Medical Image Computing and Computer Assisted Intervention--MICCAI 2018: 21st International Conference, Granada, Spain, September 16-20, 2018, Proceedings, Part II 11}, pages 210--218. Springer, 2018.

\bibitem{wang2023avalon}
Shenzhi Wang, Chang Liu, Zilong Zheng, Siyuan Qi, Shuo Chen, Qisen Yang, Andrew Zhao, Chaofei Wang, Shiji Song, and Gao Huang.
\newblock Avalon's game of thoughts: Battle against deception through recursive contemplation.
\newblock {\em arXiv preprint arXiv:2310.01320}, 2023.

\bibitem{wang2022medclip}
Zifeng Wang, Zhenbang Wu, Dinesh Agarwal, and Jimeng Sun.
\newblock Medclip: Contrastive learning from unpaired medical images and text.
\newblock {\em arXiv preprint arXiv:2210.10163}, 2022.

\bibitem{xiao2023simulating}
Bushi Xiao, Ziyuan Yin, and Zixuan Shan.
\newblock Simulating public administration crisis: A novel generative agent-based simulation system to lower technology barriers in social science research.
\newblock {\em arXiv preprint arXiv:2311.06957}, 2023.

\bibitem{xu2023language}
Zelai Xu, Chao Yu, Fei Fang, Yu~Wang, and Yi~Wu.
\newblock Language agents with reinforcement learning for strategic play in the werewolf game.
\newblock {\em arXiv preprint arXiv:2310.18940}, 2023.

\bibitem{yuan2024mora}
Zhengqing Yuan, Ruoxi Chen, Zhaoxu Li, Haolong Jia, Lifang He, Chi Wang, and Lichao Sun.
\newblock Mora: Enabling generalist video generation via a multi-agent framework.
\newblock {\em arXiv preprint arXiv:2403.13248}, 2024.

\bibitem{zhang2023biomedclip}
Sheng Zhang, Yanbo Xu, Naoto Usuyama, Hanwen Xu, Jaspreet Bagga, Robert Tinn, Sam Preston, Rajesh Rao, Mu~Wei, Naveen Valluri, et~al.
\newblock Biomedclip: a multimodal biomedical foundation model pretrained from fifteen million scientific image-text pairs.
\newblock {\em arXiv preprint arXiv:2303.00915}, 2023.

\bibitem{zhang2023attention}
Yunlong Zhang, Honglin Li, Yuxuan Sun, Sunyi Zheng, Chenglu Zhu, and Lin Yang.
\newblock Attention-challenging multiple instance learning for whole slide image classification.
\newblock {\em arXiv preprint arXiv:2311.07125}, 2023.

\bibitem{zhou2022learning}
Kaiyang Zhou, Jingkang Yang, Chen~Change Loy, and Ziwei Liu.
\newblock Learning to prompt for vision-language models.
\newblock {\em International Journal of Computer Vision}, 130(9):2337--2348, 2022.

\end{thebibliography}
